\documentclass[conference,10pt,letter]{IEEEtran}
\usepackage{graphicx}
\usepackage{array}
\newcommand{\PreserveBackslash}[1]{\let\temp=\\#1\let\\=\temp}
\newcolumntype{C}[1]{>{\PreserveBackslash\centering}p{#1}}
\newcolumntype{R}[1]{>{\PreserveBackslash\raggedleft}p{#1}}
\newcolumntype{L}[1]{>{\PreserveBackslash\raggedright}p{#1}}

\usepackage{cite}

\ifCLASSINFOpdf
\else
\fi
\usepackage{url}

\usepackage{hyperref}

\begin{document}
%
\title{Evaluation of Smartphone IMUs for\\ Small Mobile Search and Rescue Robots}

\author{
\IEEEauthorblockN{Xiangyang Zhi}
\IEEEauthorblockA{School of Information\\ Science and Technology\\
ShanghaiTech University\\
Shanghai, China 200031\\
Email: zhixy@shanghaitech.edu.cn}
\and
\IEEEauthorblockN{Qingwen Xu}
\IEEEauthorblockA{School of Information\\ Science and Technology\\
ShanghaiTech University\\
Shanghai, China 200031\\
Email: xuqw@shanghaitech.edu.cn}
\and
\IEEEauthorblockN{S\"oren Schwertfeger}
\IEEEauthorblockA{School of Information\\ Science and Technology\\
ShanghaiTech University\\
Shanghai, China 200031\\
Email: soerensch@shanghaitech.edu.cn}
}


%


\maketitle
\begin{abstract}
Small mobile robots are an important class of Search and Rescue Robots. Integrating all required components into such small robots is a difficult engineering task. Smartphones have already been made small, lightweight and cheap by the industry and are thus an excellent candidate as main controller for such robots. In this paper we outline how ROS can be used on Android devices and then evaluate one sensor which is very important for mobile robots: the Inertial Measurement Unit (IMU). Experiments are performed under static and dynamic conditions to measure the error of the IMUs of three smartphones and three professional IMUs. In the experiments we make use of a tracking system and an autonomous mobile robot.
\end{abstract}


%
\IEEEpeerreviewmaketitle

\section{Introduction}
Safety, Security and Rescue Robotics (SSRR) includes a wide area of applications that require many different kinds of robots, for example also flying \cite{MurphyHurricanes05,BirkJIRS2011} and marine vehicles \cite{murphy2011use, Rathnam2011CoordExploration2D}. Ground robots are of particular interest for many different tasks like automated transportation and exploration or remote manipulation in dangerous areas. 

One important task is the search for victims in rubble piles, which requires very small and mobile robots. Snake robots are a popular choice for this task \cite{miller200213, konyo2008ciliary}. A mechanically more simple approach is to use tracked robots for propulsion. But quite often safety, security and rescue robots are big, because the researchers try to integrate advanced sensors, ample computation, motors and batteries, for example \cite{SSRR07-JacobsAutonomy, Kang05}. This can also be observed with the robots typically participating at the RoboCup Rescue League \cite{sheh2012advancing, Sheh11}. 

We propose to use very small and cheap tracked robots for search of victims in rubble piles, similar to the prototype depicted in Figure \ref{fig:smartphoneRobot}. One main challenge in designing small robots is to integrate all require hardware like batteries, sensors, computation and communication into a small, lightweight and cheap package. Fortunately the consumer industry already spend billions of USD on developing and optimizing such devices - in the form of smart phones. 

Those devices integrate a multitude of components that a robotics engineer otherwise would have to add to the robot himself:

\begin{figure}[tb]
\centering
\includegraphics[width=1\linewidth]{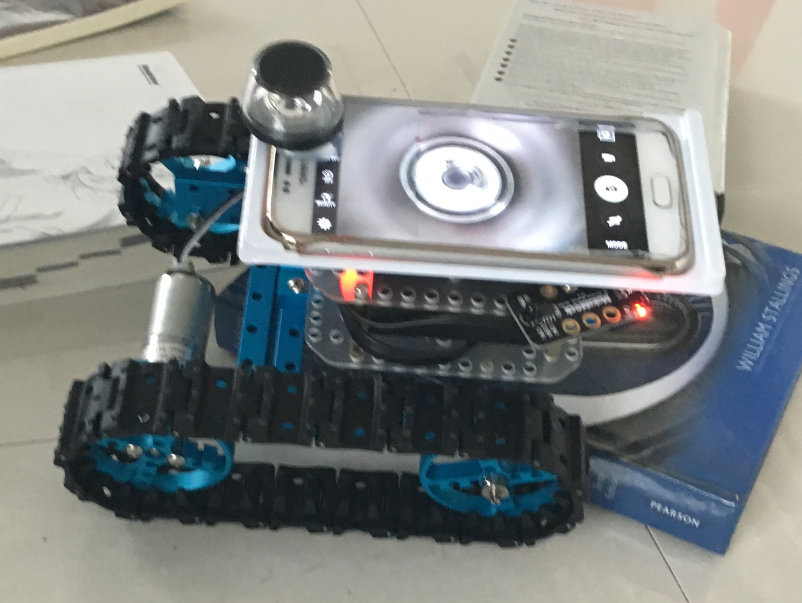}
\caption{Example of a basic tracked robot controlled by a smartphone.}
\label{fig:smartphoneRobot}
\end{figure}

\begin{itemize}
 \item Powerful computation in the form of multi-core CPUs and GPUs.
 \item Plenty of storage to save sensor data for later reports.
 \item A multitude of communication channels like 4G, WiFi, Bluetooth and NFC, which each might be used for communication with control stations or other robots but also for other purposes like localization and victim search.
 \item A battery with very good battery management.
 \item GPS receivers, including dGPS capabilities.
 \item Microphones and speakers, possibly for communication with victims.
 \item Cameras for localization, mapping and object recognition.
 \item IMUs for localization.
 \item A screen for human robot interaction - where the human might be the robot operator or a victim.
 \item A light source (screen, flash) - an essential utility for search in rubble piles.
 \item An often water- and shock-proof housing.
\end{itemize}

Basically there are very few capabilities of the smartphone that can be considered useless for robotics. This might be the case with the vibration actuator or the finger-print sensor. 

The "only" other capability that a smartphone enabled robot has to provide is locomotion, which will typically require tracks, motors, motor drivers and drive batteries. The motor drivers will be connected to the smart phone by its USB port, potentially requiring an additional micro controller like an Arduino board.

Of course we are not the first to propose the usage of smartphones for robotics. For example research has been done on smartphone based robotics for students 
\cite{oros2013smartphone} 
\cite{kasyanik2014low}
\cite{araujo2015integrating} 
 \cite{lopez2016andruino}. 

The camera and the IMU of the phones has also been used for visual odometry
\cite{tomavzivc2015fusion} 
and for more general research on small robots
\cite{aroca2012towards} 
\cite{barbosa2015ros}. 

Another interesting development is the utilization of smartphones for flying vehicles
\cite{loianno2015flying} 
\cite{aldrovandi2015smartphone}. 

The usage of smartphones in the area of search and rescue robotics has already been proposed in 2013
\cite{seljanko2013low} 
and 2014 \cite{luo2014smartphone}. 
In 2015 a full-day ICRA workshop on Robotics and Automation Technologies for Humanitarian Applications ("Where we are and where we can be") was held, where also smartphones were discussed \cite{madhavan2015robotics}. 

There is also the notable initiative of the Open Academic Robot Kit. It is designed to use 3D printed robots for lowering the barrier of entry for research into response robotics \cite{sheh2014open}. The usage of smartphones as the main control unit for such robots is a natural extension of this idea.

Localization is a crucial part for search in rubble piles. We plan to use the omnidirectional camera visible in Figure \ref{fig:smartphoneRobot} for visual localization. But that might be very challenging given the tight environment and the typically very dusty and thus uniformly colored surroundings. That is why we also have to utilize the Inertial Measurement Unit (IMU) for localization. But the question is how well the IMUs in the smartphones work, compared to expensive industry grade IMUs. In this paper we thus evaluate the quality of a number of smartphone IMUs and compare them with professional and expensive Xsense MTi IMUs. We use a 3D tracking system with 12 cameras (Naturalpoint Optitrack Prime 13) to gather the ground truth data. 

There have been previous reports on utilizing the IMUs of smartphones for robotics 
\cite{nam2013smartphone}, 
especially also for Simultaneous Localization and Mapping (SLAM)\cite{faragher2012opportunistic}. Another aspect is the calibration of the IMUs 
\cite{tedaldi2014robust}. 

The paper is structured as follows: Section \ref{sec:rosAndroid} describes how the Robot Operating System (ROS) can be used on smartphones. 
The experiments performed with the IMUs are introduced in Section \ref{sec:experiments} and 
Section \ref{sec:results} analyses and discusses the results. The paper finishes with the conclusions in Section \ref{sec:conclusions}.

\section{ROS on Android}
\label{sec:rosAndroid}

There are two methods to run ROS on Android, one is using the Android NDK and the other is using ROSJava. 

The Native Development Kit (NDK) for Android is a set of official tools that allow to use C and C++ code in Android applications, and developers can exploit it to build from their own native code or to make use of prebuilt libraries. With Android NDK, we can cross compile ROS packages and then run them on Android, and here\footnote{\url{http://wiki.ros.org/android_ndk}} is a detailed tutorial about how to cross compile ROS packages with Android NDK and several examples. This method is very suitable for ROS developers who are used to C++, i.e., Android NDK makes it convenient to reuse existing C or C++ libraries. However, as Android is based on Java, it often increases additional complexity to use NDK than Java, of course, efficiency is improved, either.
 
ROSJava\footnote{\url{https://github.com/rosjava}}, developed by Google, is a pure Java implementation of ROS, and it implements roscore and interface with ROS topics, services and parameters. ROSJava is the base of ROS on Android. Here\footnote{\url{http://rosjava.github.io/rosjava_core/latest/index.html}} is the ROSJava official on-line document. In addition, there is an Android core library in ROSJava, which is used to develop ROS applications on Android. Because the Android core library is developed in the gradle-android studio environment, it is very convenient to develop custom ROS applications on Android. As an electronic product powered by Android, we can exploit all its sensors, such as camera and IMU, and transfer sensor data on ROS, without developing any hardware drivers. In addition, we can even deploy external sensors or devices through the USB port if the product supports USB Host.

In this paper, we use ROSJava to obtain the IMU data of the smart phones. The main reason is that our application is light-weight and without much computation, so it is more appropriate to develop with Java.

In the Android core library, the RosActivity class who extends the  Activity class is the base class. In the RosActivity class, the following three steps will be done: Firstly, start the NodeMainExecutorService as a service. Secondly, run the MasterChooser activity to ask the user to connect to a remote master or create a master on this Android device. Thirdly, show a notification informing the user that ROS nodes are running in the background and the user can shut it down by clicking the notification. Normally, we can develop custom applications by extending RosActivity, besides, there are several useful layouts to be used in Android core library, such as RosCameraPreviewView, RosTextView and so on. Moreover, some good examples are also given, varying from pure software applications to hardware applications. Here\footnote{\url{http://rosjava.github.io/android_core/latest/index.html}} is the Android core library official on-line document.

It's remarkable that, in ROS, two nodes cannot share the same name, however, ROSJava doesn't help to make sure the nodes on it are unique like other ROS client libraries, so the programmers should take care. One recommended method is to add the device's IMEI number to the end of the ROS nodes' name if the same ROS node is running in more than one Android device and they are managed by the same ROS master.

In the Android API, there are not only drivers for real sensors, such as gyroscope and accelerometer, but also virtual sensors, such as rotation vector sensor and gravity sensor, which are generated by fusing other real sensors' raw data with algorithm. Rotation vector sensor combines data of accelerometer, gyroscope and geomagnetic sensor and outputs the orientation of the device. Because of the usage of geomagnetic sensor, its coordinate system has the following characteristics:
\begin{itemize}
	\item X is tangential to the ground and points approximately East.
	\item Y is tangential to the ground and points toward the geomagnetic North Pole.
	\item Z points toward the sky and is perpendicular to the ground plane.
\end{itemize}

In our experiment, we use rotation vector sensor data as the orientation of the phones, as it contains less noise and drift than using only gyroscope or accelerometer. We develop an Android application based on ROSJava. Firstly, we extend AbstractNodeMain class to create a node to publish orientation and set the node name and topic name to IMEI number of the phone to avoid conflict with the same application running on other phones. Secondly, we extend RosActivity class as the main activity, then we make sure that whether the device really has the sensor, if not, prompt the user that the sensor is absent, otherwise, we register a listener to the sensor data. Thirdly, we define a node instance to publish orientation when the listener hears the sensor data. 

\section{Experiments for the IMUs}
\label{sec:experiments}
\subsection{Multiple Sensors}
Phones of different types and professional IMU in different modes are tested in this work. In order to find the difference between cheap and expensive 
mobile phones, the cheaper one is the "m1 note" from Meizu whose price is about $\$150$ while the another two are Samsung Galaxy S6 (price about $\$650$). MTi-30 AHRS from Xsens 
is token as professional IMU in this test (price about $\$3,000$). MTi-30 AHRS use XKF3i algorithm to fusion the acceleration and the magnetic field to get the orientation. 
In different applications, XKF3i algorithm use different filter parameter to make MTi-30 AHRS work better. In other words, users should configure appropriate 
filter profile by hand. Five available filters for MTi-30 AHRS are general, high\_mag\_dep, dynamic, low\_mag\_dep and VRU\_general respectively. 
Considering their application in our research and the lack of machines, general, dynamic and VRU\_general are tested in the experiment. Respectively, they 
are applied in the case of camera tracking, mounting on person and car.

\subsection{Setup}
\begin{figure}[tb]
\centering
\includegraphics[width=1\linewidth]{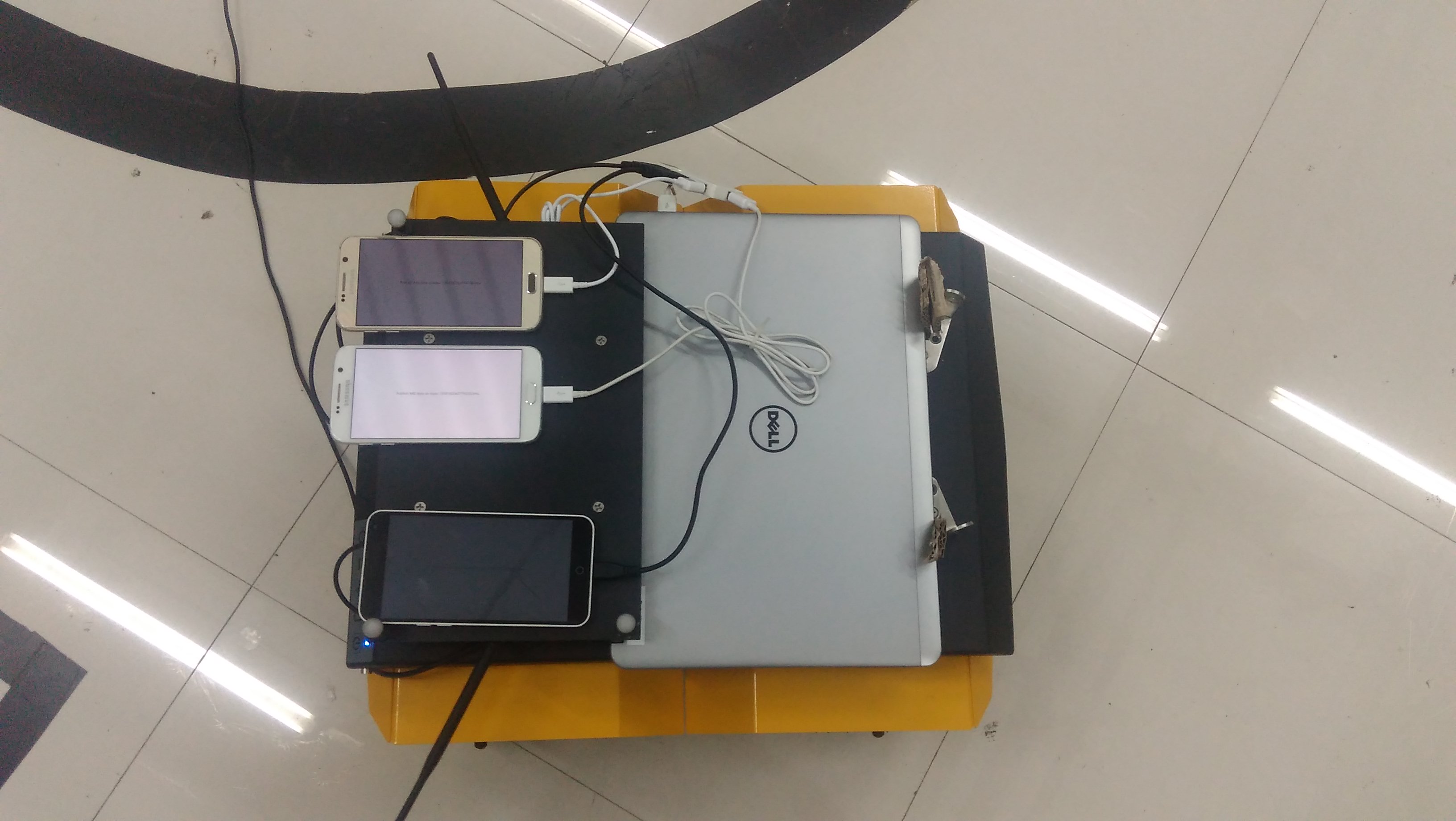}
\caption{The mobile robot (Clearpath Jackal) used for the experiments. The three smartphones and the data collection laptop can be seen - the Xsens MTi are "below deck".}
\label{fig:mti}
\end{figure}

In order to give consideration into the request of both static and dynamic test, all machines are mounted on a Clearpath JACKAL robot as Figure \ref{fig:mti} shows. 
The three phones are put on the top of the robot and the three MTi-30 AHRS are put on the next floor of the robot, which can not been seen from the figure. 
In this case, the robot only plays a role of the platform in static test. In dynamic test, the robot will drive around. Details about the test will be introduced
the following sections. Besides, there are three markers which are located in the corner of the top floor of the robot. 
In the dynamic test, these markers are used for opti tracking system to determine the pose of robot.

\begin{figure*}[tb]
\centering
\includegraphics[width=1\linewidth]{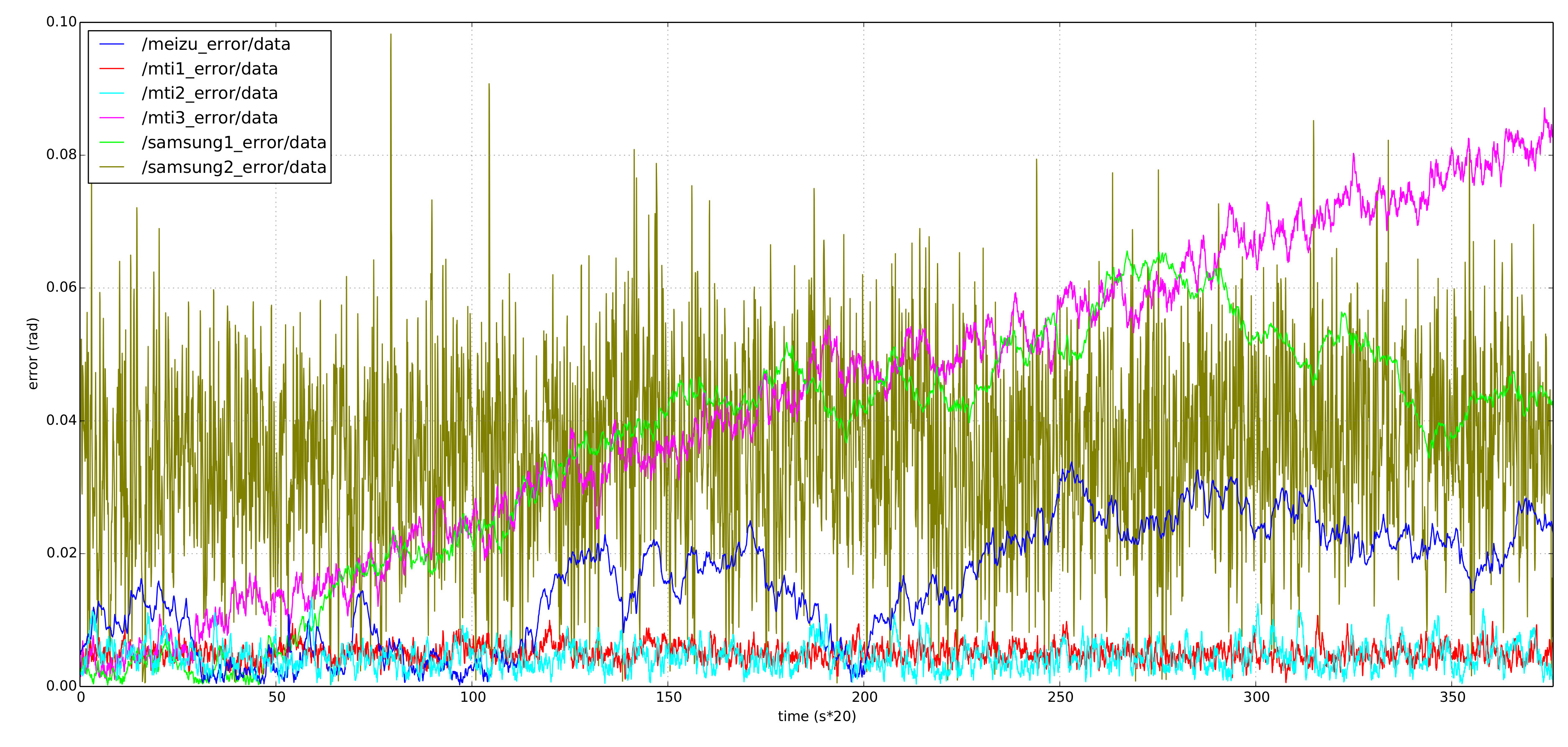}
\caption{Results of the static IMU test.}
\label{fig:static}
\end{figure*}

\subsection{IMU Static Test}
During the static test, the robot mounted all IMUs. The machines are kept stationary for more than an hour to collect the raw data. The orientation of the IMU in each sensor was 
gathered in the test. In order to decrease the time of handling data, the data was down-sampled to one message per second for each sensor. The experiment is 
designed to figure out the drift of IMU as time goes on so that let the initial state of each machine is the reference coordinate. The mean of first thirty messages 
is consider as the initial state to reduce random error. That is to say, the mean of quaternion that embodies orientation should be calculated. However, 
it is difficult to solve the problem especially when the number of samples are large. Alternatively, quaternion was transformed to RPY (roll, pitch, yaw) and then 
the mean of the RPY was computed as the reference. Then the comparison among the drift at different time could be done. In addition, the drift of different machines 
or the same machine in different modes should also be compared to find the difference.

\subsection{IMU Dynamic Tests}
Compared with the static test, the pose from opti tracking system is be added into the comparison, which is used as the ground truth for evaluating the sensors. However, a problem is that each IMU may have its own initial coordinate system. Though the first few messages play the role of the reference that can decrease the error, the case can not be handled that there may exist a skip from $-\pi$ to $\pi$. In detail, there is a range when using quaternion or RPY to represent orientation. In omni scene, the range is usually from $-\pi$ to $\pi$ in one axis. That is to say, there is a position where $-\pi$ and $\pi$ is nearly overlap. In order to handle this case, an appropriate initial state is chosen such that there is not a hop nearby.
In addition, the initial state of each machine is adjusted in similar direction, which can be considered a rough calibration. 

\subsubsection{Dynamic Test I}
In the first dynamic test, the robot is driven continuously for about one and half an hours. To make the robot drive as in the real environment, we use a remote wireless controller to make the robot drive autonomously in a random fashion. All pose data from machines and tracking system is recorded during the driving. Similar to the static test, the data was down sampled and all quaternion messages are transformed to RPY messages. Then the messages about the drift relative to the reference are published and the drift is calculated as Equation \ref{eq.1} shows. As we can predict, the drift changes very sharply when the robot is moving. In this situation, it is difficult to compare the mobile drift. Therefore, the goal of the test is to compare the drift in the beginning and at the end.

\begin{equation} \label{eq.1}
drift = \sqrt{roll^2+pitch^2+yaw^2}
\end{equation}

\subsubsection{Dynamic Test II}
The second dynamic test improves the design of dynamic test I. According to the first dynamic test, most of the data is bot useful because the data is rambling when the robot is mobile. In addition, there is only one group of data, the ending, that represents the drift, which may increase the random error.
In order to avoid the shortcoming of the first dynamic test, the process of dynamic test II is as follows:
\begin{itemize}
\item Step 1: Keep the robot static and record the orientation of each sensor for about one minute.
\item Step 2: Drive the robot for about ten minutes.
\item Step 3: Collect the data as described in Step 1.
\item Step 4: Repeat Step 2 and Step 3 for several loops.
\end{itemize}
In this test, set the number of loops is eight. For each interval, the mean drift is calculated, such that the difference among different intervals could be compared. Besides, the drift here is different from static test. Taking the advantage of rough calibration and planar motion, the drift can be embodied by the error around one axis. In other words, we only need to calculat yaw for each machine. Regarding the tracking system, the rotation axis is $y$ so that the drift is pitch. 

\section{Results and Analysis}
\label{sec:results}
After collecting and handling data, the diagram was drawn that shows the relation of drift corresponding to time among different machines. It is worth mentioning that MTi1 is in dynamic mode, MTi2 is in general mode and MTi3 is in vru\_general mode, as described in the previous section.
\subsection{Results of Static Test}
In the static test, all the IMUs are kept static without any external force. Figure \ref{fig:static} shows the result of this test. Firstly, samsung2 shows an abnormal result compared to the other devices. Too much noise causes a lot difficulties to compare samsung2 with others. Therefore, we would only compare the other sensors. In general, phones show a moderate performance while the performance of MTi is diverge. The drift of MTi1 and MTi2 is very small and does not have large fluctuation while that of MTi3 is continuously ascending. In other words, professional IMU in vru\_general mode does not have a good static performance, which is much worse than MTi in dynamic and general mode. Besides, the drift of the meizu phone is smaller than that of samsung phone. 

\subsection{Results of Dynamic Test}
\subsubsection{Continuously moving}
\begin{figure}[tb]
\centering
\includegraphics[width=1\linewidth]{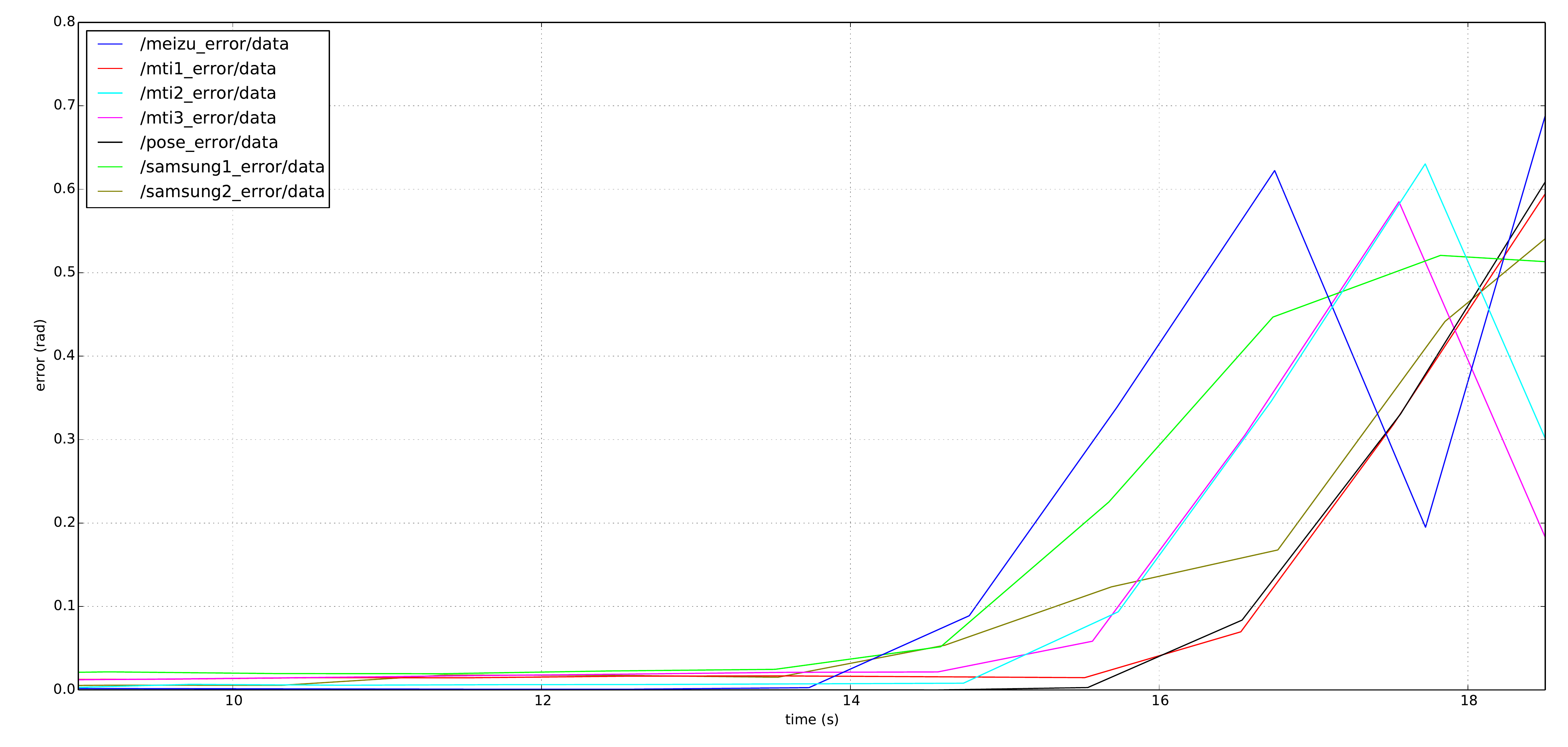}
\caption{Results of the continuous dynamic IMU test.}
\label{fig:dynamic1}
\end{figure}

In this test, the original goal that comparing the start and end data was not achieved because it is difficult to find the end data among lots of drastic changes in the data. In order to find the difference of error between moving and static, our focus turns to the interval from the beginning till the robot is moving, as shown in Figure \ref{fig:dynamic1}. Most devices, except the two samsung phones, show the same variation trend as vrpn$\_$pose from the tracking system. In addition, MTi1 is nearly synchronous with tracking system. However, each machine show a big difference of error between the static and mobile state. Besides, Figure \ref{fig:dynamic1} reflects another case that some messages are ahead of the message from tracking system, which may be caused by asynchronous situation when sending and receiving messages. Meanwhile, these asynchronous problem increase the difficulty of comparison.

\subsubsection{Alternately moving}
\begin{figure}[tb]
	\centering
	\includegraphics[width=1\linewidth]{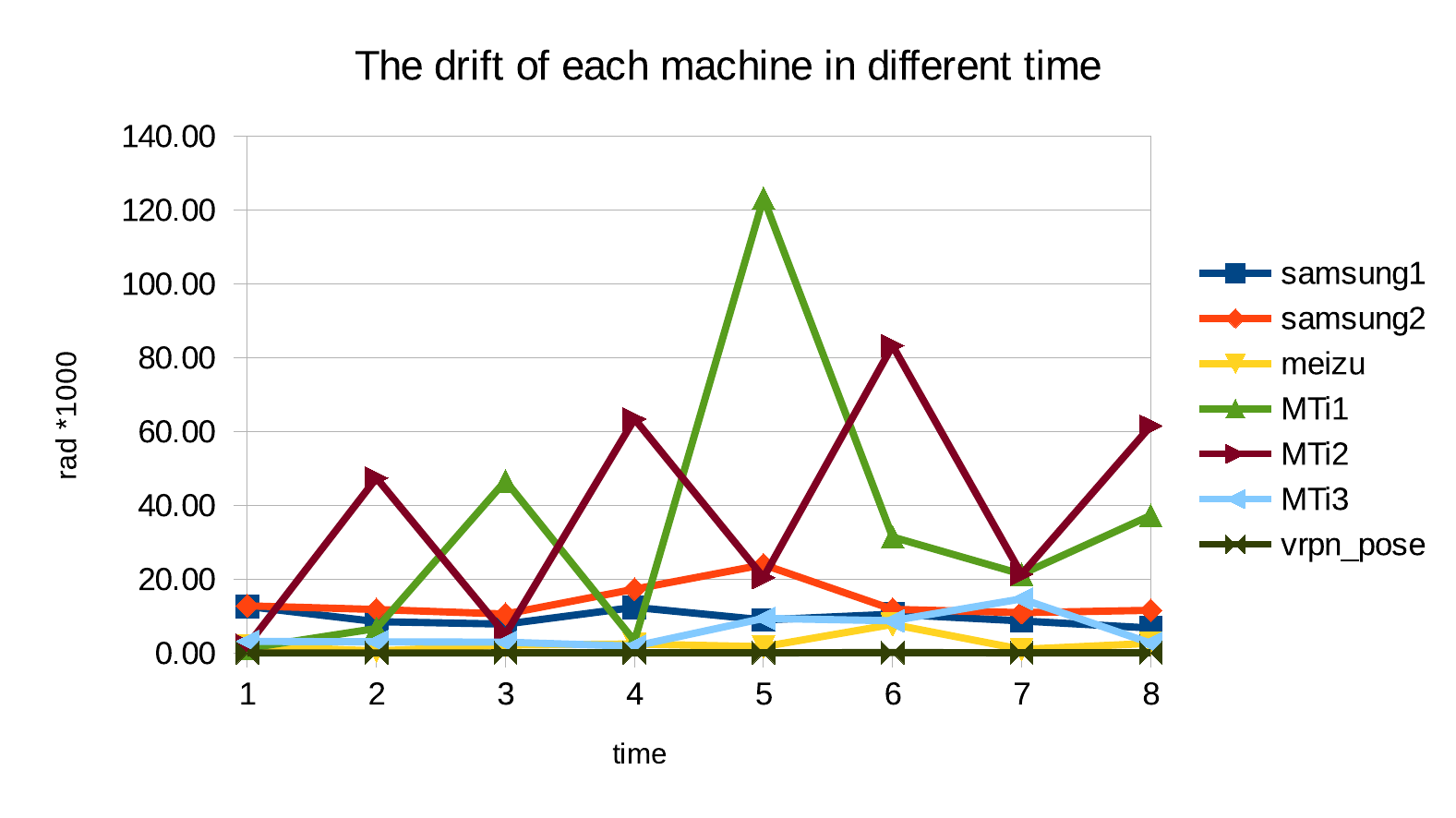}
	\caption{Results of the alternative dynamic IMU test.}
	\label{fig:dynamic3}
\end{figure}

The alternately mobile test makes up for the disadvantages of continuously mobile test. It uses the mean to solve the synchronous problem. The mean error in each interval is shown in Table \ref{dynamic2}. Firstly, the error from the optitrack tracking system (vrpn$\_$pose) is so small that it can be considered as the accurate value. Secondly, there are some larger values in the table which indicates that some sensors do not perform well when the robot is moving. In order to compare the drift of each sensor more easy, a chart generated from Table \ref{dynamic2} is shown in Figure \ref{fig:dynamic3}. We could easily find that the three mobile phones and the MTi3 have a great performance in the dynamic test while the drift of MTi1 and MTi2 is much larger. 

Comparing dynamic test and static test, it can be easily found that different modes of MTi-30 AHRS have a balance between static and dynamic state. Mti has a better performance in static test when it is in dynamic or general mode. Meanwhile, it performs better in mobile test when vru$\_$general filter is used. Regarding the mobile phones, the samsung2 shows a abnormal performance and another two phones have a performance between the worst and the best case of MTi-30. 
\begin{table}[tb]
\centering
\caption{\label{dynamic2} Results of the alternative dynamic IMU test $(rad*10^{3})$.}
\begin{tabular}{|c|c|c|c|c|}
\hline
Sensor & $t=0$ & $t=1$ & $t=2$ & $t=3$ \\
\hline
Samsung1 & 12.63 & 8.51 & 7.83 & 12.39 \\
\hline
Samsung2 & 12.78 & 11.79 & 10.57 & 17.31 \\
\hline
Meizu & 1.96 & 0.55 & 2.16 & 2.44 \\
\hline
MTi1 & 1.27 & 6.66 & 46.65 & 3.26 \\
\hline
MTi2 & 2.04 & 47.41 & 5.05 & 63.39 \\
\hline
MTi3 & 3.16 & 3.04 & 2.94 & 1.86 \\
\hline
Vrpn$\_$pose & 0.02 & 0.02 & 0.03 & 0.03 \\
\hline
\hline
Sensor & $t=4$ & $t=5$ & $t=6$ & $t=7$ \\
\hline
Samsung1 & 8.96 & 10.52 & 8.70 & 6.76 \\
\hline
Samsung2 & 23.93 & 11.84 & 10.93 & 11.58 \\
\hline
Meizu & 1.71 & 7.77 & 0.99 & 2.63 \\
\hline
MTi1 & 123.24 & 31.43 & 21.37 & 37.28 \\
\hline
MTi2 & 20.48 & 83.28 & 21.34 & 61.50 \\
\hline
MTi3 & 9.39 & 8.73 & 14.66 & 2.66 \\
\hline
Vrpn$\_$pose & 0.03 & 0.13 & 0.04 & 0.04 \\
\hline
\end{tabular}
\end{table}

\section{Conclusions}
\label{sec:conclusions}
In this paper, we estimated and compared the drift of Inertial Measurement Units (IMUs) in static and dynamic experiments. We configured three Xsens MTi-30 in the three different modes: general, dynamic and VRU\_general. We also used three smartphones: one Meizu m1 note and two Samsung Galaxy S6. The static experiment indicated that the MTi-30 in general and dynamic mode greatly outperform the rest of the IMUs, and that the drift of MTi-30 in VRU\_general will raise with time, which makes its performance the worst. The reason for this behavior is that in this mode the yaw from the compass is ignored because it is deemed to be too noisy. In the dynamic experiment, we regarded the orientation obtained by tracking system as ground truth and compared it with that of IMUs. In this case, the MTi-30 in general and dynamic mode perform worst while MTi-30 in VRU\_general performs much better. It is remarkable that the drift of Meizu is least, even though it is the cheapest among these IMUs. The comparison of the performance of the MTi-30 in different modes indicated that there is a balance during different circumstances, we should configure MTi-30 to appropriate modes according to our purpose.

As the experiments have shown we can see that smartphone IMUs offer performance not too worse when compared to professional IMUs. We also see some differences between the smartphone IMUs. Thus more research and evaluations of specific IMUs have to be performed before selecting the best smartphone (IMU) for use in small mobile robots.


%
%

\IEEEtriggeratref{14}


%


\bibliographystyle{IEEEtran}

\bibliography{references,References_2,References_old}             

\end{document}